\documentclass{article} 
\usepackage{iclr2023_conference,times}

\usepackage{amsmath,amsfonts,bm}

\def\eqref#1{equation~\ref{#1}}

\def\1{\bm{1}}

\DeclareMathAlphabet{\mathsfit}{\encodingdefault}{\sfdefault}{m}{sl}
\SetMathAlphabet{\mathsfit}{bold}{\encodingdefault}{\sfdefault}{bx}{n}

\usepackage{hyperref}
\usepackage{url}
\usepackage{ulem}
\usepackage{comment}
\usepackage{amsmath}

\title{ChatGPT is all you need to decolonize \\
sub-Saharan vocational education}

\author{Isidora Tourni$^{*}$\\
Boston University\\
\texttt{isidora@bu.edu} \\
\And
Georgios Grigorakis$^{*}$ \\
National and Kapodistrian University of Athens \\
\texttt{grigorakis5@gmail.com}
\And
Isidoros Marougkas$^{*}$\\
Rutgers, the State University of New Jersey\\
\texttt{im316@scarletmail.rutgers.edu}
\AND
Konstantinos Dafnis\\
Rutgers, the State University of New Jersey\\
\texttt{kd703@cs.rutgers.edu}
\And
 Vassiliki Tassopoulou  \\
University of Pennsylvania \\
\texttt{vtass@seas.upenn.edu}
}

\iclrfinalcopy 
\begin{document}
\maketitle
\begin{abstract}
\vspace{-5pt}
The advances of Generative AI models with interactive capabilities over the past few years offer unique opportunities for socioeconomic mobility. Their potential for scalability, accessibility, affordability, personalizing and convenience sets a first-class opportunity for poverty-stricken countries to adapt and modernize their educational order. As a result, this position paper makes the case for an educational policy framework that would succeed in this transformation by prioritizing vocational and technical training over academic education in sub-Saharan African countries. We highlight substantial applications of Large Language Models, tailor-made to their respective cultural background(s) and needs, that would reinforce their systemic decolonization. Lastly, we provide specific historical examples of diverse states successfully implementing such policies in the elementary steps of their socioeconomic transformation, in order to corroborate our proposal to sub-Saharan African countries to follow their lead. 
\end{abstract}

\vspace{-10pt}
\section{Introduction}
\vspace{-7pt}
The incoming fourth industrial revolution, as expressed in terms of the recent achievements and investments in designing and implementing Artificial Intelligence (AI) systems, promises to disrupt the socioeconomic world order. Data-driven platforms that manipulate multimedia and interact with humans in real time offer the opportunity to automate processes and increase productivity. They lower the barrier of technical knowledge required by the end user for performing specific tasks, and are capable of running in a distributed and personalized manner - directed by their interaction and fit to their respective needs - as well as reducing the economic cost per capita. 

Large Language Models (LLMs) have recently emerged as the most celebrated example of such systems, with the recent launching of ChatGPT taking over the Internet by storm. LLMs are AI models that use
Deep Learning to process and generate natural language
text. These highly overparameterized models are trained on massive amounts of text data, allowing them to learn the nuances and complexities of human language. One of the key advantages of these models is their ability
to understand the context of a given prompt and generate appropriate responses. This is a significant improvement over earlier language models,
which were often unable to interpret the meaning and intent behind a given piece of text. With its ability to draw out knowledge, answer insightful questions and teach complex matters, it is obvious that education is the first field to be disrupted.  Additionally, other types of generative models, that have met success, such as Dall-E-2 \citep{Dalle2}, Stable Diffusion \citep{Stable_Diffusion}, Jukebox \citep{Jukebox} and Phenaki \citep{Phenaki} can be incorporated in applications based on LLMs in order to provide a novel, holistic learning experience. 

Undoubtedly, the countries that can mostly benefit from such a drastic alteration of the status quo are the developing ones, which are frequently plagued by scarcity of resources. The automation of labour and the democratization of expertise paves the way for rapid growth and improvement of the living standards of their citizens, should they incorporate such models in their educational systems successfully.

As a matter of fact, the mere characteristics of those AI systems tend to favour the Technical and Vocational Educational Training (TVET) in comparison to traditional academic trajectories. One main reason for that is the fact that they offer the advantage of immediate application of the acquired knowledge on the field and the flexibility to apply informed feedback for correcting mistakes made either by the end user or the AI. A second one is their innate relationship with the imminent needs of people in developing countries. As they often face challenges to fulfill 
fundamental needs, the practical effects of TVET can show up faster and produce palpable life improvements. The distributed nature of TVET, which can be efficiently applied to educational setups with minimum equipment, that teach a variety of non time-critical subjects, is a great fit 
for the expertise provided by LLMs 
that can often be improved via interaction.

However, it is common for African countries to allocate their resources towards establishing populated, monolithic universities while neglecting TVET. This obstructs the application of the "Diffusion of Innovations" theory \citep{Diffusion_of_Innovatopm}, that has proven to be the most effective policy direction amongst newly-founded states. As highlighted by researchers with expertise in the liberated African educational models \citep{Teferra}, \citep{Mohamedbai}, the establishing of traditional top-down academic institutions in decolonized countries has been plagued by lack of funding, adequate expert staffing, academic freedom and gender balance \citep{Teferra}. The common factor that emerges from all the aforementioned cases is the remnants of the relationship of those states with their colonial past. As they often have inherited parts of the systems of their former Central European colonizers - nations with completely different background, mentality and evolutionary motivations - they possess the collective tendency to mimic Anglo-Saxonic educational standards and express social favoritism towards academic distinction means for acquiring status. This is a model that evidently does not fit developing countries, as it assumes the existence of well-functioning institutions and well-founded legal ordinance as prerequisites, a domain in which they fall behind, according to a variety of indices \citep{Index1}, \citep{Index2}. So,  
in reality, these educational environments slowly adopt technological innovations 
and provide obsolete knowledge with only implicit application to real life. These discrepancies have Africa consistently remain 
on the dark side of every technological revolution. Indicatively, sub-Saharan Africa accounts for just 1.06$\%$ of the world’s total AI publications, while, on the other side, reports of traumatic experiences felt by low-paid Kenyan workers during data-labelling and quality assessing tasks of internet content for detecting and removing toxicity has seen wide publicity \citep{Press}.

On the contrary, we claim that sub-Saharan African countries should adopt a bottom-up educational approach, that prioritizes TVET over traditional academia and puts (augmented) LLMs in its heart. Our main argument is that such a framework will provide the maximum flexibility for scaling the effect of this dynamic technology, and nourish novelty.

The main contributions of this position paper are:
\vspace{-5pt}
\begin{itemize}

    \item[$\bullet$] we provide tangible application examples of LLMs in TVET,
    
    \item[$\bullet$] we introduce historical examples of growth in developing countries that
    prioritized TVET over classical academic education

    \item[$\bullet$] we discuss the advantages that LLMs offer over traditional search engines in that specific context  
\end{itemize}
\vspace{-7pt} 
    in order to, ultimately, suggest an alternative educational framework that is based on the  hitherto achievements of LLMs and that would allow sub-Saharan countries to establish a flexible higher educational system with focus on TVET.

\vspace{-10pt} 
\section{Background}
\vspace{-7pt}
\subsection{Sub-Saharan African educational systems}
\vspace{-7pt}
Extensive research has been conducted in the individual characteristics of African higher education, highlighting both its merits and its fallacies. In \citep{africanhandbook}, K.D. Bryant  
considers the relation between colonial educational systems and the first sub-Saharan African educational systems directly proportionate. However, it is still argued that despite its Eurocentric narrative, racially biased mentality and narrow viewpoint, that system unintentionally sowed the seeds of emancipation, as most indigenous post-colonial elites found themselves in such positions after successfully utilizing  their accumulated assets for their own end. 

In \citep{ojiambo2018education}, financial deprivation and state-level frailty are described as factors detrimental for the success of post-colonial Sub-Saharan African educational systems, while some attempts of Sub-Saharan African states towards TVET practices are also described. 
\vspace{-7pt}
\subsection{TVET vs Academia}
\vspace{-7pt}
The issue of whether investment and national focus on solely tertiary education is a Eurocentric, post-colonial remnant is discussed in \citep{ojiambo2018education}, where it is argued that attempts to found a more TVET-oriented educational system in various Sub-Saharan African Nations during the decades of 1980 and 1990, were mostly unsuccessful due to a lack of a consolidated political vision-aspiration on the matter. Attempts towards that direction were spontaneous, anaemic and ill-funded, resulting in a perpetual distrust of the local populace that was never persuaded to not opt for traditional and prestigious tertiary education admission. 
Another paragon towards neglecting TVET due to colonial history, was the destruction of an inter-generational lore of local technical expertise that we analyse below as Indigenous Knowledge (IK). While academics and tenure professors of tertiary education were hardly available, indigenous TVET teachers were virtually non-existent, especially in the case of post-apartheid South Africa. \citep{RSA_TVET}
\vspace{-7pt}
\subsection{Large Language Models and Generative AI}
\vspace{-7pt}
Generative AIs can generate novel content, rather than simply manipulating existing data. They encode input information into a latent high-dimensional space and deploy generator models capable of effectively transforming  text to other modalities 
in a conversational way, answering follow-up
questions, challenging incorrect premises and rejecting inappropriate requests. It is based on a Transformer architecture \citep{BERT} and trained through Reinforcement Learning with Human Feedback \citep{RLHF}. Examples of such models are DALL-E 2 \citep{Dalle2}, which generates original realistic images and art from a text prompt, stable diffusion \citep{Stable_Diffusion}, which performs image alterations at the level of latent space of a diffusion model, Phenaki \citep{Phenaki}, which generates videos from time variable prompts, and Jukebox \citep{Jukebox}, which employs a hierarchical VQ-VAE architecture to compress audio into a discrete space.  

\vspace{-7pt}
\section{Success Stories: 3 diverse cases}
\vspace{-7pt}
With poverty, 
and high entropy being the norm for the newly founded or liberated states throughout human history, there are numerous examples that African countries can draw inspiration from in order to imitate their optimal practices. Indeed, there are well-documented historical examples of such a deliberate bottom-up approach in allocating educational resources with spectacular results. 

\vspace{-5pt}
\subsection{The case of the Greek state of 1830}
\vspace{-5pt}

Greece found itself on the paradoxical position of establishing its first modern State on two of the most destitute administrative divisions of the Ottoman Empire. One of the first questions that occurred just after the mere founding of that state, was how to structure the educational system, with the elite of the academic upper class being in favor of founding universities, in accordance to European Enlightenment standards. However, the government at the time deliberately decided to prioritize TVET: 
one example was farming education, based on the methods developed by the Swiss agronomists and educationalists Philipp Emanuel von Fellenberg and Johann Heinrich Pestalozzi \citep{Kleoniki}.  
The results of this approach vindicated the Greek government then.

A hundred and thirty years after Kapodistrias' reforms, by 1960, the formerly backwards Greek nation had a much higher GDP per capita than Spain and Portugal, two former colonial Great Powers \citep{Statistic}. By 2008, it was rendered the state with the highest GDP per capita among all that liberated themselves from the Ottoman Empire, and was a member of some of the most exclusive state-"clubs" in the world, namely the European Union, NATO, the OECD etc., sporting one of the largest numbers of higher education graduates per capita in the Globe. The experience stemming from the aforementioned case leaves no doubt that emphasis on vocational and technical skills may prove astonishingly fruitful in creating a stepping stone of growth for recently decolonized societies nowadays. It further implies that a top-down approach of tertiary education as the pinnacle of a nation's progress is not an unshakable belief.

\vspace{-5pt}
\subsection{The case of the South Korean state of 1960s-1970s}
\vspace{-5pt}

In the occasion of South Korea, vocational training was top-down prioritized on a State level during the 1960s and 1970s under the concept of fostering development of the then-impoverished nation (basic Law for Vocational Training in 1976 and  previously the Vocational Training Act of 1967). The Korean economic boom of the late 1980s is at least partially attributed to the resulting well qualified industrial worker cohorts stemming from the aforementioned vocational and technical system. \citep{South_Korea}. By almost tripling TVET attendance and offered positions, the nascent Korean industry, which was producing only labour-intensive goods and was under an immense shortage of relevant manpower, managed to fill up its ranks by 1980. The subsequent economic miracle of a 100-fold increase in national income per capita, in a matter of 30 years, is attributable to the quick and efficient creation of a large industrial worker base that propelled the Korean industry to progressively becoming one of the global hotpoints of production. Lastly, the indigenization of the whole TVET procedure (handbooks, terminology, curriculum) by 1980 was equally important in helping form a distinct Korean industrial and working culture.\citep{korea2nd}

\vspace{-5pt}
\subsection{The case of Singapore of 1960s}
\vspace{-5pt}

At the same period, though under different conditions, is the case of Singapore. In the late 1960's, under the imminent threat of large-scale unemployment as a result of the British withdrawal, Singapore overtook society's initial reluctance towards blue-collar jobs in favor of academic education. With the intention to brace the drive towards industrialisation, the government introduced measures that included the extension of technical training to the primary and  secondary level with mandatory application to the entire male and half the female cohort. This revised common lower-secondary curriculum aimed to introduce students to manual skills, to cultivate interest and reinforce practical culture. In 1968, 73$\%$ of primary school leavers went to the academic stream, while just 12$\%$ went to the technical stream and 15$\%$ to the vocational stream, respectively. Likewise, in secondary schools, 92$\%$ of graduating students took the academic route, while only 2$\%$ went to the vocational stream and 6$\%$ to the technical stream. As a result of those efforts, in less than 5 years, there was a 300$\%$ increase of interest in TVET studies \citep{Singapore1}, \citep{Singapore2}, \citep{Singapore3}.

\vspace{-7pt}
\section{Potential Applications}
\vspace{-10pt}
Applications of our proposal are bipronged: On the one hand, such an option may be combined with other media, in a small amount of time and on a wide, simultaneous dissemination among interested individuals; skipping all logistical hassle such endeavours may introduce when the human factor is instrumental at all stages of a given procedure. On the other hand, our approach allows for a complete alteration of the existing interface 
towards other disciplines and/or professions. 
\vspace{-7pt}
\subsection{Agricultural and Food Sector}
\vspace{-7pt}
A LLM finetuned for introducing Agricultural and Nutritional TVET combined with elements of IK best practices could alleviate colonially-entrenched malfunctions via a modernization of suboptimal practices. Similarly, such models may swiftly be altered towards providing other AI-generated mini courses targeted to knowledge- and capital- unintensive occupations, even with Augmented Reality and NERFs interfering in the procedure. In a nutshell, such a tool will efficiently and with low cost educate massive cohorts of students, promote indigenization and total decolonization and, above all, provide the means for a quick leap forward in economic terms, similar to the one we previously witnessed in Korea - but this time with minimized risks and maximized quality, quantity and general benefits.

\vspace{-7pt}
\subsection{Collaborative Robotics (Cobots)}
\vspace{-7pt}
As Africa is rapidly transforming into the outsourced manufacturing center both for the West, as well as the booming Upper East, familiarization with the practical interfaces of intelligent "co-bots" emerges as a necessity. The West’s "algorithmic invasion" that "simultaneously impoverishes development of local products while also leaving the continent dependent on its software and infrastructure" \citep{Birhane} that already happened in the software industry (e.g. Nigeria, one of the more technologically advanced countries in Africa, imports 90$\%$ of its software) should not be expanded to the catholic inclusion of robotics in working spaces. 

Here, (enhanced) LLMs may play the role of privileged technical advisors, 
in the specialization of workers and relevant technicians, especially since modern robotics are turning themselves towards LLMs for creating their own user-interfaces \citep{Liang}.

\vspace{-7pt}
\subsection{Paramedics and First-aid Education}
\vspace{-7pt}
Lastly, an area in which such a technology could immediately be useful is in building a quick and effective first-response line to emergency medical situations, such as epidemics and accidents caused by natural disasters, on a community level, complementary to existing systems \citep{slingers2022ten,Nigerias_first_response}. Generative models could prove to be invaluable tools in properly and quickly training Emergency Medical Service Workers to help in emergency massive vaccinations, provide first-aid services, patient transportation, treatment of effects of traumatic events, and other types of rapid care and support to their communities. The advantage of training local people to handle such a response lies in that they have a solid knowledge of the community structure, problems, practices and resources, and hence can focus on learning and adapting techniques best fit for the region's needs. People with a light or no medical background could be quickly trained using the platform to be employed in handling emergency situations and avoid further expansion of major health crises and events. The increased level of volunteerism in those areas constitutes an advantage when considering the number of people who would be willing to undergo such a training, by simply offering their time and deep understanding of community demands, which could prove life-saving given that often there are few or no trained practitioners there ready to provide such a response.

\vspace{-7pt}
\section{Language models vs Traditional Search Engines}
\vspace{-7pt}

One question that may naturally arise is what is really so different and revolutionary in incorporating LLMs in educational systems, compared to traditional search engines that have been around for decades now.
 Our suggestion is that LLMs may ultimately carve the path towards a complete decolonization of Sub-Saharan education. As literature already states, the modus operandi of the former colonizers was preserved in their reluctant withdrawal, in the form of norms, institutions and social structures of post-colonialist nature, under emergency circumstances. Teaching of what is described as Indigenous Knowledge (IK) \citep{IKorismos} under the newly independent Nations was initially non-existent. Various attempts during past decades to include IK in the educational curriculum of Tanzania, Kenya, Nigeria and other educational systems, were met with a severe lack of funding, trained manpower and social trust on the initiatives. \citep{IKSYRIZA} LLMs can bridge that gap, by being fine-tuned on the dynamic and constant inclusion of IK elements.
 
 Instead of only providing generic functionalist results into indigenous students and trainees, as a conventional search engine does, a LLM may additionally provide a genuinely comprehensive, inclusive and more akin to sustainability set of results through the incorporation of IK. Also, 
 multi-modal systems, based on the interconnection and cross-feedback of the assorted media-specialized AIs, set the foundation for educational tools that far exceed the capabilities of search engines. 

\vspace{-10pt}
\section{Limitations}
\vspace{-10pt}
Of course, similarly to every software technological advancement, 
hardware remains a principal limitation. As a matter of fact, focusing on TVET, instead of on higher-level education, may lower the initial investment cost barriers of founding steadfast educational institutions, but access to high-resolution and low-latency delivery mechanisms, as well as reliable, high-throughput internet connectivity and electricity rise as bare necessities, now more important than ever. Moreover, no matter the improvements that every successive version of ever-growing models brings to the quality of their output, both the abstract nature of human communication as well as training resources limitation are not expected to allow them to produce flawless results. So, a need for high-order supervision conducted by human experts with educational role still remains, even if it is sparser than today. Last but not least, the argument of the previous section could be inverted. With few training data that consider African cultural and economic realities, the output of ChatGPT-style generative AIs could be skewed toward reinforcing colonial ideological hegemony.

\vspace{-10pt}
\section{Conclusion}
\vspace{-10pt}
We reckon that prioritizing the inclusion of LLMs in sub-Saharan African educational systems will be paramount step towards their decolonization. In that effort, TVET rises as the most promising path to shift resources to, and will immediately unveil the beneficial effects of these novel technological features, 
as has been demonstrated by assorted historical evidence.

\newpage

\bibliographystyle{iclr2023_conference}
\bibliography{iclr2023_conference} 

\end{document}